\documentclass{article}
\usepackage{graphicx}
\usepackage[a4paper, total={7in, 10in}]{geometry}
\usepackage{multicol}
\usepackage[sorting=none]{biblatex} %
\title{Enhancing Balance and Posture in Material Handling Tasks with a Wearable Robotic Tail}

\author{Eisa Anwar(1), Sajeeva Abeywardena(2), Stuart Miller(3), Ildar Farkhatdinov(4)}

\begin{document}

\maketitle

(1)Faculty of Natural, Mathematical and Engineering Sciences, King's College London, London, UK. This work was conducted whilst at: School of Electronic Engineering and Computer Science, Queen Mary University of London, London, UK.

(2)School of Mechanical Engineering Sciences, University of Surrey, UK.

(3)William Harvey Research Institute, Queen Mary University of London, London, UK.

(4)School of Biomedical Engineering and Imaging Sciences, King's College London, London, UK

\section*{Abstract}

Frequent heavy lifting is linked to musculoskeletal strain, with lower back pain affecting up to 40\% of manual labourers. Inspired by animals using tails for balance, we developed a wearable robotic tail as a dynamic counterbalance. Ten participants performed an imitated shelf loading/unloading task across three conditions: active tail, passive tail, and no tail. Kinematics, centre of pressure, and muscle activation (gastrocnemius, erector spinae, tibialis anterior, biceps brachii) were measured. Compared with no tail, the active tail achieved $>$20~mm reduction in centre of pressure peak displacement, 4.3$^\circ$ reduction in trunk motion, and $>$9\% and $>$20\% reductions in erector spinae and gastrocnemius activity, respectively. The passive tail produced smaller reductions, suggesting passive counterbalancing helps while active movement augments it further. Biceps brachii activation increased with the active versus passive tail. The ankle kinematics and tibialis anterior activation were unchanged. These findings support wearable robotic tails for improving balance during material handling.

\section*{Introduction}

Across the animal kingdom, tails serve as versatile organs for balance, locomotion, and stability. Kangaroos use their muscular tails as functional fifth limbs during locomotion, providing both propulsive force and postural support \cite{10.1098/rsbl.2014.0381}. Cheetahs exploit tail inertia to execute rapid aerial reorientation during high-speed pursuit \cite{Wilson2013Cheetah}. Lizards use their tails for mid-air righting reflexes, maintaining stability during falls and jumps \cite{Libby2012LizardTail}. These biological examples demonstrate a fundamental biomechanical principle: an articulated appendage, by shifting the body's centre of mass (CoM), can dynamically counteract destabilising forces without the need for ground contact adjustments. Despite the elegance and ubiquity of this strategy in nature, it remains largely unexplored as a design principle for wearable assistive robotics.

Many industries involve manual handling, where workers are required to perform repetitive physical load-carrying tasks, which are closely linked with the development of musculoskeletal (MSK) injuries. Examples include assembly lines \cite{OTTO2017467}, construction \cite{ANTWIAFARI201741}, vehicle maintenance \cite{Sampouw_2019} and aircraft maintenance \cite{doi:10.10520/EJC169017}. In logistics, repetitive lifting tasks are particularly prevalent, involving delivery couriers \cite{Ryu2023}, warehouse workers \cite{logistics5040089, Waters01121998, Kamat_2017}, and baggage handlers \cite{doi:10.3233/WOR-172494, 10.1093/annhyg/mew043}. Furthermore, the increase in delivery services during the COVID-19 pandemic has increased the demand for these roles \cite{Bogue2022}.

The repetitive nature of these tasks has been shown to fatigue muscles, particularly in the lower back, and increase spinal bending, which is a well-documented risk factor that increases the risk of MSK strains and injuries \cite{DOLAN1998713, s22155655, GALLAGHER201584, 94882163025bc558d9a66736cf04d8af12dfa064, Russeng_Saleh_Wahyulianti_Palutturi_2021}. Some muscles have a heightened risk, with the lower erector spinae (ES) fatiguing at a higher rate \cite{ANTWIAFARI201741}, and increasing the risk of developing lower back pain \cite{su14105767}, with up to 40\% of manual labour workers in industry, agriculture and construction reporting pain  conditions \cite{2202794}. Assistive robotics technology can improve working conditions and minimise risk of developing health conditions.

Lifting tasks shift the body's Centre of Mass (CoM) away from the middle of the body, moving it closer to the limits of the Base of Support (BoS). The BoS refers to the area within the contact points of the body and the environment (e.g. for a standing person, it includes the space covered by and between their feet). When the horizontal position of the CoM, equivalent to the Centre of Pressure (CoP), moves outside the BoS, the body becomes unstable and will fall unless stabilised. To maintain stability, individuals can modify their posture to recentre their CoP, e.g. by leaning \cite{jacobson2021balance} or adjust their BoS, e.g. by taking a step. BoS adjustments can be difficult as space is not always available, with many industrial tasks restricting the ability to adjust footing. For example, workers loading shelves in front of them may be restricted in repositioning their footing anteriorly. Instead, they must rely on postural adjustments and CoM movements to maintain stability, which can place additional stress on the body, for example by employing a forward bending strategy at the hips. Tasks such as these that require reaching or extending the arms can increase the load on the lower back and further increase the risk of injury \cite{Ryu2023}.

The growing adoption of robotics and automation mitigates these risks by enabling more tasks to be performed by robotic systems, which can enhance productivity and safety. Examples include robotic arms \cite{sym13020226} and autonomous mobile pallet vehicles \cite{dhaliwal2020warehouse, Bogue2022, laber2020impact}. Despite these advancements, a substantial portion of industrial tasks remain reliant on human labour. Several factors contribute to this: the high initial costs of robotic solutions can make them cost-prohibitive for many smaller businesses or those in low- and middle-income countries (LMICs). Additionally, humans excel in tasks requiring adaptability and decision-making in complex or variable environments, where robots may struggle \cite{7136589, https://doi.org/10.1111/j.1467-8667.2004.00365.x}. Consequently, many workers remain in roles that demand repetitive manual handling tasks that risk their safety and long term health.

To address these challenges, various wearable robotic systems have been developed to provide physical assistance. In a prior review~\cite{10762871}, we explored the scope of technology capable of supporting balance and posture. We found a variety of technologies, including exoskeletons, support harnesses, supernumerary robotic limbs (SRLs), mobile robot canes, and wearable gyroscopes~\cite{10762871}. Exoskeletons represented the most studied, offering the highest improvements in postural support and reduced muscle strain \cite{10.3389/frobt.2018.00053, mi10070463, ijerph20156468, ijerph18052677}.

However, exoskeletons have notable limitations. They are often expensive \cite{s22249987}, time consuming to setup and wear \cite{martinez2024}, and require precise alignment with the user's joints for effective operation \cite{Lindhard2023}. Misalignment could cause discomfort or even injury during use. Furthermore, exoskeletons can restrict the Range of Motion (RoM) of the user, especially in complex or confined spaces, such as vehicle loading / unloading, which can lead to discomfort \cite{BALTRUSCH201894, 10.3389/fpubh.2022.1039680, Schwerha03042022, 00043764-202203000-00004}. Moreover, they are typically task specific, making them less adaptable to varied workflows \cite{Siedl14122023, 10.3389/fpubh.2022.1039680}. These factors have limited their widespread adoption and retention in industrial settings.

An emerging alternative to exoskeletons are SRLs. Unlike exoskeletons, which enhance existing joints, SRLs augment the body with additional limbs, providing extra movement capabilities. These devices can extend the BoS by interacting with the environment \cite{6907002, 7139896}, or adjust the CoM to support the load and mitigate against destabilising forces \cite{10719748, 9341323, 10.1145/3306214.3338573}. SRLs offer several advantages. They are not reliant on precise alignment with the user's joints, making them more adaptable to varied tasks as the bodies joints are free to move naturally. This allows for quicker donning and doffing (i.e. putting the system on and taking the system off), making them more suitable for workers who may regularly alter their tasks, such as delivery drivers switching between driving and manual handling. The lower complexities of these systems also reduce the cost of these systems and the need for extensive training and maintenance.

Until now, SRLs have not yet been evaluated in load manipulation tasks that analyse their biomechanical and physiological effects on the body, which is an essential step in evaluating their potential to alleviate body strain. Previously, researchers have proposed using supernumerary robotic tails to explore immersive human-machine interaction \cite{10.1145/3311823.3311847} and demonstrate that tail-like wearable devices can affect human balance~\cite{9341323} including our previous modelling studies~\cite{abeywardena2022human,abeywardena2023towards,abeywardena2024mechanical,abeywardena2024modelling}, but to date no research has provided an experimental evidence on the use of wearable robotic tail-like devices to support posture and balance during material handling.

\begin{figure}
    \centering
    \includegraphics[width=0.9\textwidth]{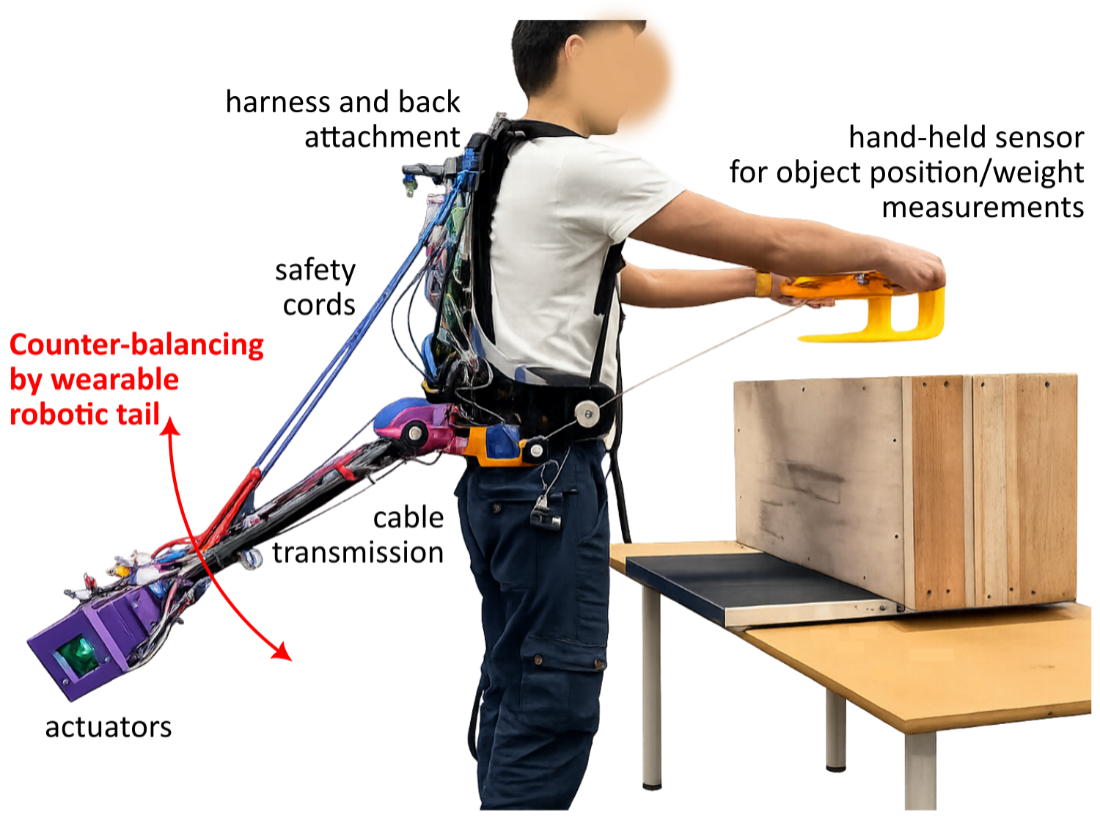}
    \caption{\textbf{Robotic tail} The custom built wearable robot tail worn by a user during object handling task.}
    \label{fig:tail}
\end{figure}

Therefore, in this study, we tested an SRL in the form of a wearable robotic tail, shown in Figure~\ref{fig:tail} that can provide physical support for material handling tasks. This work builds on our previous research, which detailed the design and development of the wearable robotic tail~\cite{10719748}. Here, we build upon that work by conducting a broad participant study that involves load manipulation tasks. To the best of our knowledge this is the first study that analyses CoP, body kinematics, and muscle activity to understand the effect of a wearable robotic tail on a user's body. We evaluated conditions with the tail in active and passive states, as well as without it to assess the impact of the device, including its active control systems, and its static effect.

\section*{Results}
\subsection*{Centre of pressure}

\begin{figure}
    \centering
    \includegraphics[width=0.9\textwidth]{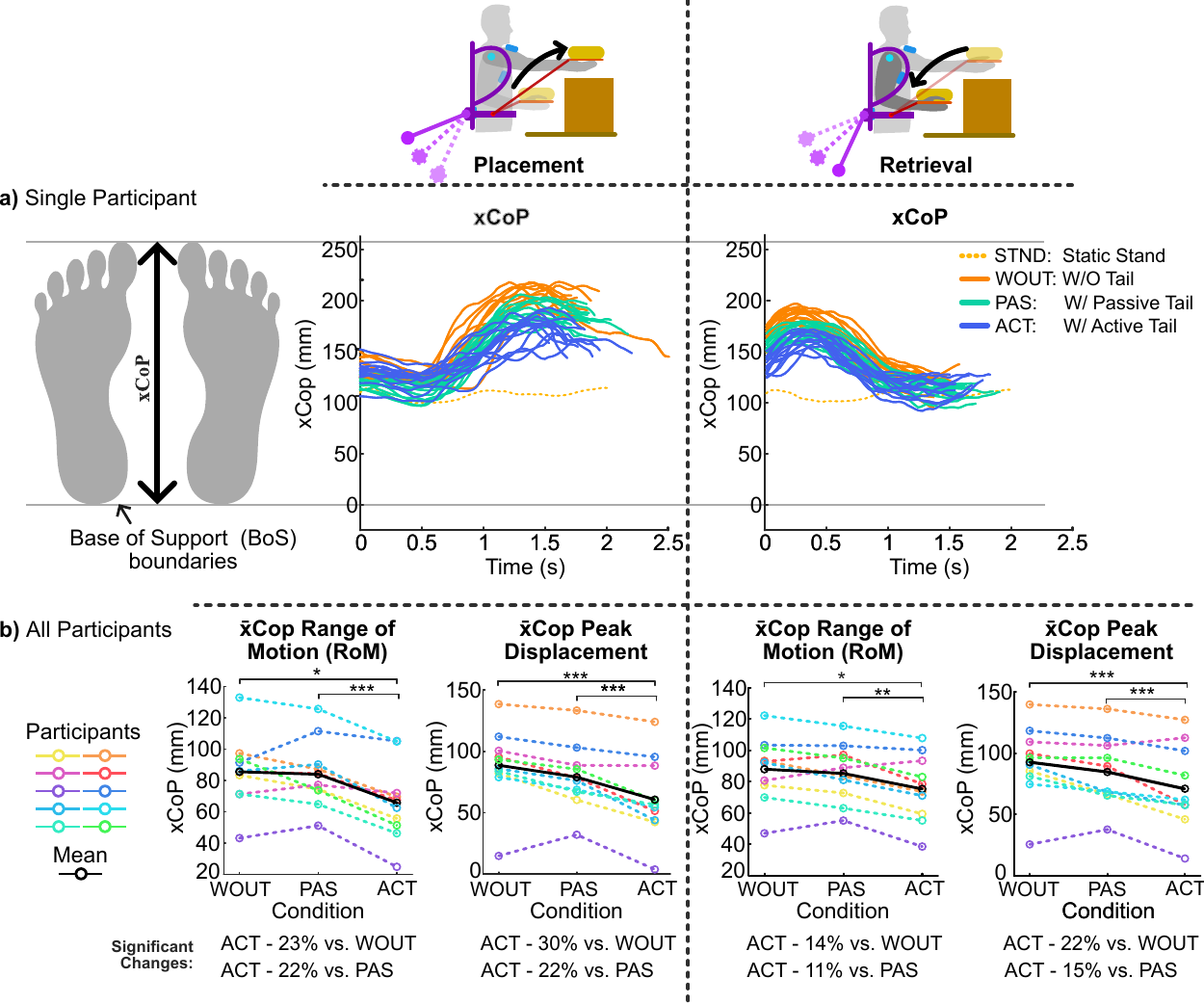}
    \caption{\textbf{Centre of Pressure (CoP) Analysis.} (\textbf{A}) CoP change for one participant for the placement and retrieval trials for all conditions of the experiment. The blue, green, and orange lines represent the active tail (ACT), passive tail (PAS), and without tail (WOUT) conditions respectively, while the static standing (STND) trial is shown in yellow. The grey lines at the top and bottom of the figure mark the boundaries of the BoS (i.e., the front and back of the feet). (\textbf{B}) Peak displacement and RoM of the sagittal CoP for all participants. Each participant is represented by a single colour, and each point represents the average of all their repetitions for a single condition. The percentage changes of the statistically significant results are stated, and asterisks indicate the significance level: *($p<0.05$), **($p<0.01$), ***($p<0.001$).}
    \label{fig:CoP}
\end{figure}

Ten participants performed load placement and retrieval tasks in three conditions, with the robotic tail (ACT), without it (WOUT), and with the robotic tail in passive mode (PAS). For clarity of explanation we present time series of the measurements for an individual participant first, followed by a summary of findings across all participants.

Figure~\ref{fig:CoP}A illustrates the sagittal CoP position and its movement across repetitions during load placement and retrieval for a single participant. In all conditions, the CoP position moves forward and backward during the object placement and retrieval tasks, respectively.  CoP position remains within the base of support (BoS) indicating stable posture and balance. For this participant, the CoP shifts substantially forward in WOUT, moving close to the front of the base of support, compared to PAS and ACT when the CoP remains closer to the centre of the BoS.

Across all participants, during placement tasks, CoP peak displacement (Figure~\ref{fig:CoP}B) was less in ACT compared to both PAS (26\,mm [30\%] reduction; $p < 0.001$) and WOUT (17\,mm [22\%] reduction; $p = 0.0014$). Similarly, total CoP's RoM was less in ACT compared to both PAS and WOUT by 18\,mm (22\%, $p < 0.001$) and 20\,mm (23\%, $p = 0.01$), respectively. No differences were found between PAS and WOUT ($p > 0.05$ for both CoP peak displacement and RoM).

For load retrieval tasks, similar reductions in CoP peak displacement and CoP's RoM were observed. With ACT, CoP peak displacement was reduced compared to WOUT (20\,mm reduction [22\%], $p < 0.01$) and PAS (13\,mm reduction [15\%], $p < 0.01$). CoP's RoM with ACT was also less compared to WOUT (19\,mm reduction [14\%], $p = 0.01$) and PAS (18\,mm reduction [11\%], $p < 0.001$). No significant differences were found between the PAS and WOUT conditions in peak displacement or range of motion for placement or retrieval tasks. This suggests that the observed support comes from the active control movement of the tail.

\subsection*{Trunk and ankle range of motion}

\begin{figure}%
    \centering
\includegraphics[width=0.9\textwidth]{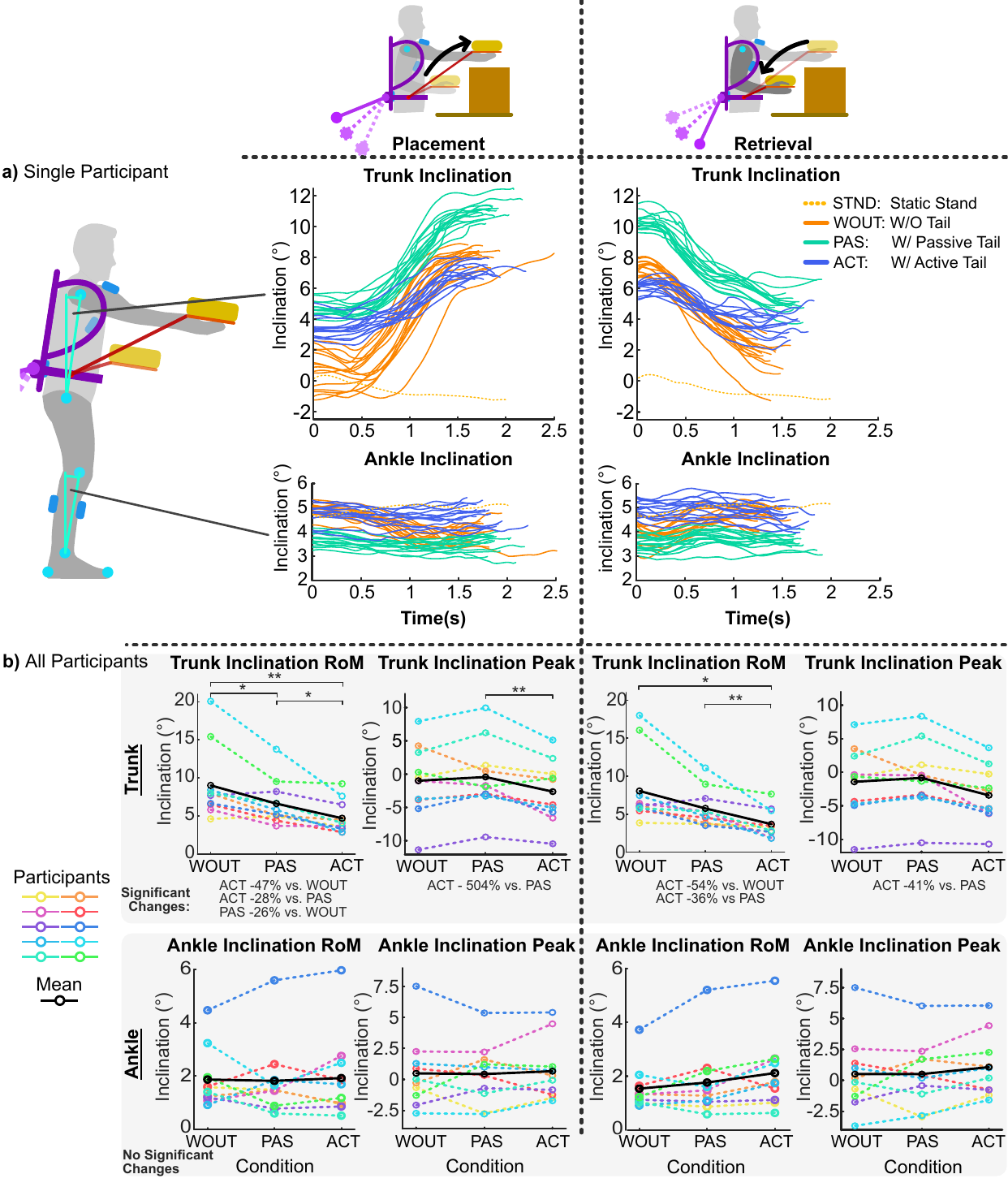}
    \caption{\textbf{Trunk and ankle inclination (i.e. flexion/extension).} (\textbf{A}) Single participant for the placement and retrieval trials for all conditions of the experiment. The single yellow line represents the static standing baseline result and the three other colours with multiple repetitions represent the three different conditions. (\textbf{B}) Trunk and ankle inclination for all participants. Each participant is represented by a single colour, and each point represents the average of all their repetitions for a single condition. Dotted lines link a single participant across their experimental conditions. The percentage values of the statistically significant results are shown under respective graphs.}
    \label{fig:Angles}
\end{figure}

Figure~\ref{fig:Angles}A shows trunk and ankle inclination across repetitions for a single participant (the same as in Figure~\ref{fig:CoP}A).
During load placement in WOUT, the trunk angle initially starts close to the static standing baseline before increasing rapidly toward the end of the repetition, indicating a progressive forward lean. In PAS, the trunk angle begins slightly elevated above the WOUT baseline, before increasing and ultimately reaching a greater peak than WOUT. In ACT, the starting trunk angle falls between the WOUT and PAS conditions but increases more gradually throughout the movement, with the final angle remaining lower than in WOUT. During load retrieval, the pattern of trunk inclination mirrors that observed in the placement trials but in the opposite direction. ACT exhibits the slowest rate of angular change and the lowest overall peak, though the angle settles marginally higher than WOUT by the end of the repetition. Ankle angle remains relatively stable across both ACT and PAS conditions, with minimal change from start to end of each task.

Figure~\ref{fig:Angles}B shows the peak displacement and RoM for both the trunk and ankle inclination for all participants. Peak trunk inclination with ACT was less compared to WOUT (4.3\textdegree{} [47\%] reduction, $p < 0.01$) and PAS (1.9\textdegree{} reduction [28\%], $p < 0.05$). Peak trunk inclination in PAS was also less than in WOUT (2.4\textdegree{} reduction [26\%], $p = 0.01$). This smaller reduction in trunk's RoM observed with PAS suggests the passive weight of the tail contributes to limiting trunk inclination. Peak trunk inclination was less with ACT compared to PAS (2.2\textdegree{} reduction, $p = 0.01$), demonstrating the capability of the robotic tail in reducing the peak trunk inclination. Peak trunk inclination in ACT and PAS did not change significantly compared to WOUT, indicating that the presence of the tail (active or passive) was not significantly affecting the peak trunk inclination.

No significant differences were observed with peak or RoM of the ankle angle in placement or retrieval tasks.

\subsection*{Muscle activity on lower back and legs}

\begin{figure}
    \centering
    \includegraphics[width=0.9\textwidth]{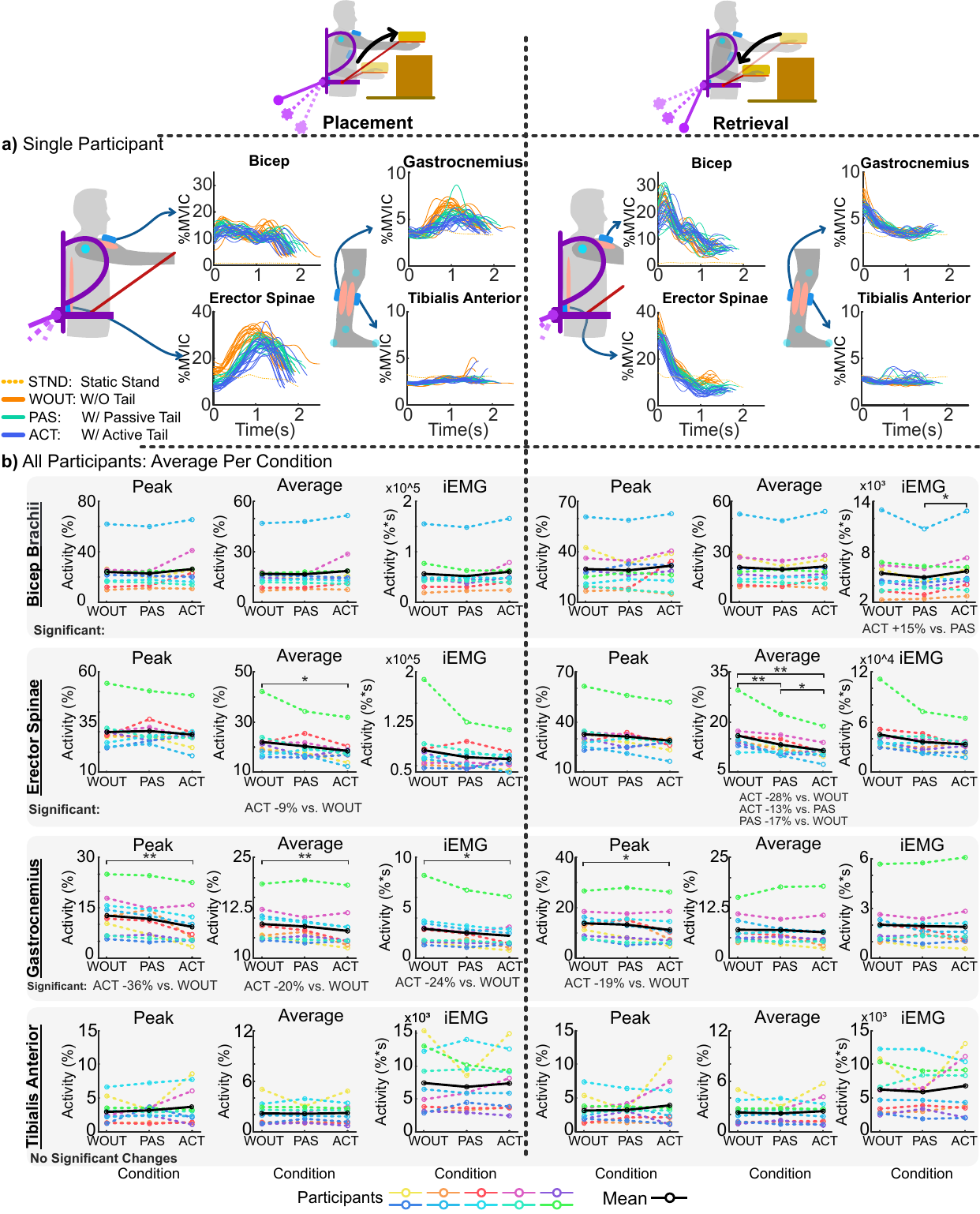}
    \caption{\textbf{Muscle Activity Data.} (\textbf{A}) Muscle activity of one participant for all conditions and repetitions. The left four graphs are for each of the four recorded muscles for the placement task and the right four graphs are for the retrieval task. (\textbf{B}) Muscle activity for all participants for all conditions. Each condition's repetitions are averaged for each individual point. The data is split into four rows with two columns. The left graphs are for placement and the right graphs are for retrieval tasks. Each row corresponds to one of the four muscles tested: biceps brachii (BB), erector spinae (ES), gastrocnemius (GS) and tibialis anterior (TA). Each section of the three graphs consists of peak activity, average activity and iEMG result.}
    \label{fig:EMG}
\end{figure}

Figure~\ref{fig:EMG}A shows the electromyography (EMG) time series for the same representative participant as in Figures~\ref{fig:CoP}A and \ref{fig:Angles}A. 
For BB, activity remained relatively constant until the final stage when the load begins to be lowered to the platform. For retrieval tasks, activity starts higher and decreases as the load is moved towards the body.

For the ES, activation increases and decreases during the placement task, with peak activation occurring approximately one second into the task. When retrieving the load, initial ES muscle activity is high, decreasing during the activity.
For the GS, activity during placement tasks are small, remaining below 10\%, with a small peak mid-movement. During load retrieval, GS muscle activity decreases as the load is moved closer to the body.
Across all conditions and movement directions, TA activity remains low (< 5\%), similar to activity recorded during the standing baseline
suggesting that for this participant wearing the robotic tail did not cause excessive TA activation.
For this participant the data shows slightly reduced activation for GS and ES in ACT condition compared to WOUT.

Figure~\ref{fig:EMG}B shows results for all participants, revealing trends in line with those observed in the single participant analysis with the exception of the biceps. For the biceps in retrieval tasks, there is a significant increase in iEMG muscle activity of 15\% in ACT compared to PAS ($p < 0.05$), but no difference to WOUT.

For ES, significant group-level changes in average activity are found for placement tasks, with 9\% lower activity with ACT compared to WOUT ($p < 0.05$). For retrieval tasks, there are also significant decreases in average activity of 28\% with ACT compared to WOUT ($p < 0.01$), and by 13\% with ACT compared to PAS ($p < 0.05$). Additionally, a 17\% reduction was observed with PAS compared to WOUT ($p < 0.01$). These findings further suggest that while PAS reduces muscle activity, ACT enhances the effect.

The GS exhibits significant reductions in activity, particularly during load placement. Peak activity is reduced by 36\% with ACT compared to WOUT ($p < 0.01$). Average muscle activity is down 20\% with ACT compared to WOUT ($p < 0.01$). iEMG is down 24\% with ACT compared to WOUT ($p < 0.05$). During retrieval tasks, the peak activity decreases by 19\% in ACT compared to WOUT ($p < 0.01$). These findings suggest better balance, as reduced GS activation indicates a lower reliance on ankle-based balance strategies, potentially reducing muscle strain and enhancing stability.

For the TA, muscle activity remained consistently low in all conditions, with no significant group-level or post-hoc changes during placement or retrieval tasks. This suggests that the tail does not overcorrect the balance or induce backward CoP shifts that would trigger TA activation, indicating that it is not overcompensating in counterbalancing.

\section*{Discussion}

Our initial expectations were that the designed wearable robotic tail will improve user’s postural stability and reduce (or at least not increase) muscle activity during material handling task, hence contributing to better ergonomics of the overall task. In terms of postural stability, the experimental study provided convincing evidence of positive robot’s contribution to the participants’ postural stability. With respect to the muscle activity, the robot did not cause the increase of muscles effort, with part of the experimental evidence showing reduction of the muscle activity for ES (lower back) and GS (ankle joint). The only exception was the increase of BB (upper limb) integrated activation in active tail condition vs. passive tail condition. In the following we discuss the results and provide our interpretation for the key findings. 

Participants’ CoP movement was used a postural stability metric which is known to provide a robust measure of overall postural stability \cite{ruhe2010testretest}. Analysis of the CoP movement showed that the active tail improved body stability by dynamically counterbalancing forward movement in the body’s CoP. The active control system of the robotic tail continuously adapted to the movement of the load carried by participants, keeping their CoP closer to the centre of the BoS. This controlled effect is also evident when looking at the inclination of the trunk, which had a lower RoM suggesting more postural stability. Active counterbalancing movements of the tail required less upper body movements of the participants as demonstrated by reduced trunk flexion during material handling, which is improving task ergonomics, reduces stress on the spine and reduces health risks~\cite{Keyserling1988TrunkPostureBackPain,Chaffin1987}. Robotic tail did not influence ankle joint movements. The participants maintained their normal foot-ankle-shank movement pattern when the robotic tail was used. Unchanged ankle joint movement with simultaneously reduced trunk movement and improved postural balance outcomes demonstrate that the robotic trail has effectively performed its counterbalancing function.

Wearable robotic tail has contributed positively to improved postural ergonomics through reducing the lower back muscle activity both when it was passive and active, with active robot achieving highest effect (ES, average EMG). Increased and prolonged ES muscle activity and higher spine loads are known to have direct links to increased lower back pain, potentially increasing the risk of future low back injury~\cite{Ferguson2005MotorRecruitmentLBP,isa2014analysis}. Hence, the use of robotic tail had a positive impact on load carrying and material handling ergonomics with potential long-term impact.

Decreased muscle activity was observed for GS muscle that primarily control ankle plantar flexion when the robotic tail was used. The counterbalancing action of the robotic tail reduced the need of higher GS activity, which is related to a reduced plantarflexion effort for postural balancing. However, the reduced GS activation was not reflected in ankle joint movements, as stated above. Simultaneous reduction in ES and GS muscles’ activity demonstrate positive effects of the robotic tail for maintaining safer and more stable upright posture during material handling tasks.

Interestingly, we have observed mixed findings on the increased BB activity (greater iEMG, but not peak or average activation) during retrieval task when robotic tail was counterbalancing compared to the condition when it was passive, although not different to when the tail was not worn. This suggests that the passive tail may reduce BB activation, but this is removed following turning the tail on. This might also be associated with the need of the arms to reach further during retrieval as the trunk moves less when the tail is used. Additionally, increased BB activity might be caused indirectly by the active robotic tails’ movement that was felt by the user through body-tail interaction forces that caused upper body muscle activity which is known to happen for other type of wearable robotic devices~\cite{Kong2022ExoskeletonEMGAnyBody}. However, as peak and average BB activation were not different in all conditions, this may be due to an increased movement duration in the active condition compared to the passive tail condition (this was not assessed within the current study).

Overall, our findings highlight the effectiveness of active counterbalancing in enhancing postural stability and reducing muscular effort, demonstrating the potential of wearable robotic tails as assistive devices for material handling. While various assistive devices are already used in industry such as passive and active exoskeletons \cite{Fox2020ExoskeletonsManufacturingPerformance}, the proposed robotic tail represents a class of SRL-based devices that have not yet been widely deployed in industrial settings. Promoting adoption will require improvements in compactness, portability, and power efficiency, which remain limitations of the current research prototype. Nevertheless, the results obtained here demonstrate that wearable artificial robotic limbs can generate beneficial physical robot–user interactions that improve postural stability and reduce muscle activity. Further technology development is needed to make such devices compact, power-efficient, and autonomous.

The use of an imitated task setup within a laboratory environment, along with the inclusion of non-professional participants, represents another limitation of this study. Such conditions do not fully capture the complexity and variability of real-world work environments. Future work should therefore focus not only on technological improvements but also on evaluating the device with a larger and more diverse cohort, particularly in real-world material handling contexts. Studies involving warehouse workers, logistics personnel, and other industrial practitioners would provide more robust evidence of the practical benefits of the wearable robotic tail and help identify usability challenges and operational constraints associated with deployment in working environments.

\section*{Methods}
\subsection*{Tail design and control}
Technical details of the robotic tail design and its modelling can be found in our previous work (43,44). The same robotic tail as described in (43) and shown in Figure 1 was used in the experiment presented in this work (also see supplementary material). 

Additionally a handheld sensor system was developed to measure the CoM offset induced by lifting a load in front of the body (shown in Figure \ref{fig:tail}). A handheld platform was built with integrated force sensors and pull rope encoders to measure the weight of a carried load placed on top of it, and its distance from the body. The total weight of the system is around 6 kg. The counterweight is 3 kg, and can extend 50 cm behind the user. A technical drawing of the robotic tail is provided in the supplementary material.

The CoP offset induced by the carried load, $\Delta xCoP_{l}$, is given by Equation~\ref{eq:CoPHumanOffset}:

\begin{equation}
    \Delta xCoP_{l} =  \frac  {D_{w} M_{w}}  {M_{b} + M_{w} + M_{t}}
    \label{eq:CoPHumanOffset}
\end{equation}
\noindent where $D_{w}$ represents the perpendicular distance between the body's CoM and the load, $M_{w}$ denotes the mass of the load, $M_{b}$ is the body mass, and $M_{t}$ is the tail mass of 3~kg. This assumes the tail mass is positioned at the body's CoM in the horizontal plane; however, in the tail's starting home position pointing vertically downward, it is actually positioned slightly posterior to the body's CoM (approximately 10~cm). As the mass of the tail is small relative to the body's mass, this small difference at the home position can be omitted.

The objective is for the tail to generate an equivalent and opposite sagittal CoP offset to that induced by the carried load. The CoP offset induced by the tail, $\Delta xCoP_{t}$, is calculated as follows in Equation~\ref{eq:CoPHumanTail}:

\begin{equation}
    \Delta xCoP_{t} =  \frac  {M_{t}L_{t}\cos{(\theta_{t})} }  {M_{b} + M_{w} + M_{t}}
    \label{eq:CoPHumanTail}
\end{equation}
\noindent where $L_{t}$ represents the tail length (50~cm) and $\theta_{t}$ is the tail angle measured from the vertical, with $0^\circ$ corresponding to a vertical downwards orientation. To balance the CoP offsets generated by the load and the tail, these two offsets are equated and solved for $\theta_{t}$, as shown in Equation~\ref{eq:CoPTailCancelHumanWeight}:

\begin{equation}
    \theta_{t} =  \cos^{-1} \left( \frac{ D_{w} \times M_{w}}{L_{t} \times M_{t}} \right).
    \label{eq:CoPTailCancelHumanWeight}
\end{equation}

\noindent The motor controllers on the tail then use this target angle, $\theta_{t}$, to regulate motor positions through a proportional-derivative (PD) control scheme which runs on a Teensy 4.1 microcontroller (PJCR, USA).

As measured in our previous validation study with the tail, the latency between the input control signal and the tail's movement is under 60~ms. This indicates that the tail responds quickly enough to deliver stabilising impulses, with a reaction time faster than the average human response time of approximately 100~ms \cite{Nashner1981}. As a result, the tail's movements should feel nearly real-time to the user, who is therefore unlikely to initiate any reactive motion before the tail has moved.

\subsection*{Experimental procedure}
We recruited 10 participants (six male, four female; ages: 20–31 years, non-vulnerable) to perform two load transfer tasks with a 4 kg load. In the first task (placement), participants moved the load from a static holding position close to their body and placed it on a raised platform in front of them. Then, after a short pause, participants picked up the load back up from the platform and brought it back (retrieval), holding it close to their body in the same static holding position. These movements aimed to simulate a shelving scenario and were performed in three conditions, with the tail (ACT), without it (WOUT), and with the tail in passive mode (PAS, in which the tail was worn but unpowered, with the tail segment resting in a vertically downward position).

The condition where the participants performed the tasks without wearing a tail (WOUT) served as a baseline. The passive mode test (PAS) was crucial for distinguishing the contributions of different aspects of the robotic tail. By comparing passive (PAS) and active (ACT) conditions, the effects of the static physical mass could be isolated from the dynamically controlled counterbalance.

\begin{figure}
    \centering
    \includegraphics[width=0.9\textwidth]{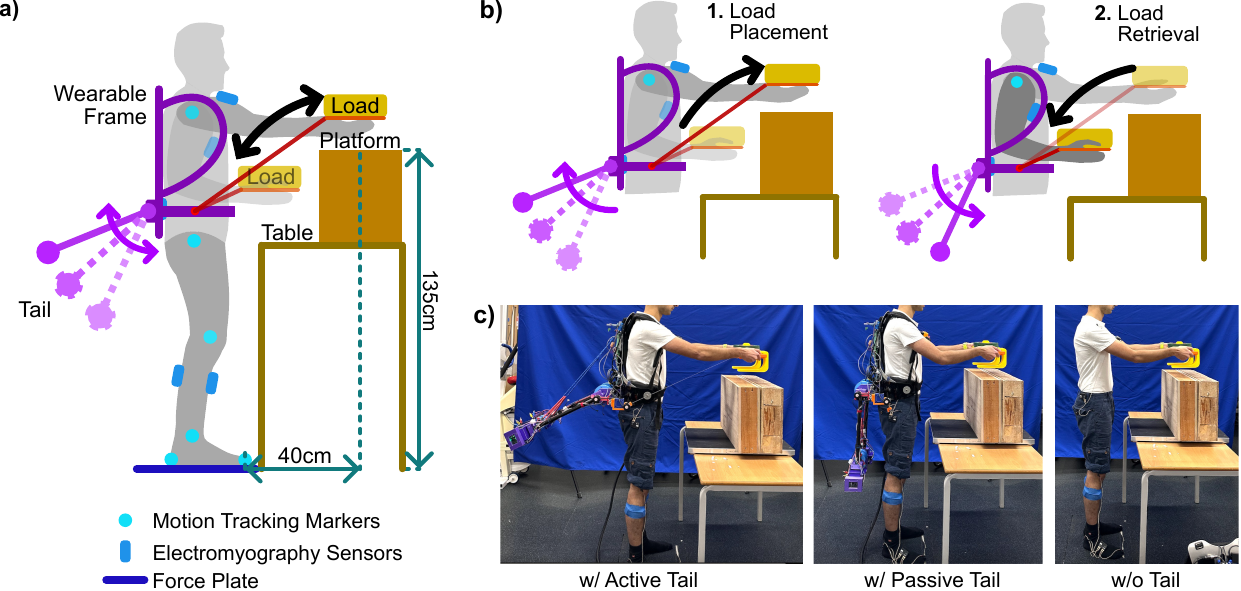}
    \caption{\textbf{Experimental Setup.} (\textbf{A}) The setup of the experiment and the sensors used. The light blue circles are motion tracking markers, the darker blue rectangles are electromyography sensors (EMG), and the dark blue line the figure is standing on is a force plate. (\textbf{B}) The experimental procedure consisting of two tasks. (\textbf{C}) The experiment was carried out in three conditions: with the tail powered and active (ACT), with the tail unpowered and passive (PAS), and without the tail (WOUT).}
    \label{fig:setup}
\end{figure}

For each condition, a participant first stood behind the table with a raised platform on top, which had a total height of 135~cm from the floor. The centre of the platform was 40~cm in front of the participant, measured from the front of their feet shown in Figure~\ref{fig:setup}A. A 4~kg load was placed on a platform in front of the participant. The load was handled using a handheld sensor. This handheld sensor is passively attached to the wearable frame of the system via a pull rope encoder and rotational encoder to measure the load's anterior distance from the body's CoM. Throughout the experiment, the participant used the handheld sensor to lift and manipulate the load.

In each condition, placement and retrieval tasks were carried out sequentially, repeated 15 times (Fig.~\ref{fig:setup}B).
Load placement trials were measured from the time the forward velocity of the load exceeded 1~cm/s, as calculated via a motion-tracking marker on the load, and ended when the load contacted the raised platform, as detected by a force plate under the platform. For load retrieval trials, repetitions started when the load was lifted off the platform (measured by the same force plate) and ended 2~cm before the load reached its closest position to the participant. Ending the trial slightly before the closest position excluded any brief pauses or adjustments that occurred when the load was held at the body, ensuring that only the active reaching movement was captured and trial lengths remained consistent. After completing each cycle, the participant briefly placed the load down in front of them before repeating the sequence to avoid over-fatiguing the muscles.

Before the experiment began, participants practised the sequence three times, with and without the tail, to familiarise themselves with the movements of the task and the feel of the tail. Furthermore, participants also completed a static standing (STND) trial to establish a baseline of an idle standing position before performing the tasks. The order of the experimental conditions was randomly assigned to each participant using a random number generator. Participants were asked to perform three practice repetitions of the task under each condition before the main experiment began, allowing them to familiarise themselves with the task and the tail. To prevent fatigue, a five minute break between conditions was provided, allowing participants to rest their muscles. With three conditions, there are six possible orders in which the experiment can be conducted. Each order was completed at least once. Ethical approval was obtained from Queen Mary University of London Research Ethics Committee (QMREC2018/48 121), informed consent was obtained after the nature and possible consequences of the studies were explained.

\subsection*{Measurements}

Multiple biomechanical and physiological metrics were monitored, including sagittal CoP, ankle and trunk inclination, as well as muscle activity. Firstly, to measure CoP, the participant performed the experiment while standing on a force plate (Kistler) with a data logging rate of 2~kHz. A fourth-order low-pass Butterworth filter with a cut-off frequency of 100~Hz was applied to the force plate data to remove high-frequency noise. We measured the peak displacement and RoM of the CoP to determine what extent the body's balance is shifted \cite{Antwi-Afari2017, Kaewmanee03092022, ijerph192416710}. The peak displacement is measured from each participant's average baseline standing CoP position (recorded for each participant during static standing without wearing the robotic tail) and the RoM is the total range of movement. We also measure how close they are to their stability limits (i.e. the distance to the front and back of their feet).

An array of position tracking markers (Codamotion) was used to record the BoS and measure ankle and trunk inclination with a sampling rate of 200~Hz. Since the task is symmetrical, markers were placed only on the right side of the body. To track the BoS, markers were placed at the first toe and calcaneus of the foot (i.e. the front and back).
As with CoP analysis, the peak displacement and RoM of these metrics are examined. The inclination of the ankle (vector between ankle and knee markers) and the trunk (vector between hip and shoulder markers) were measured relative to the average inclination recorded during the static standing trial. The ankle plays a crucial role in stabilisation and balance strategy~\cite{Lee02012015, Chow2005}, while trunk inclination is a key indicator of overall stability and strain~\cite{10322313, doi:10.3233/WOR-205067}.%

Finally, eight surface electromyography (EMG) sensors (Delsys Trigno, USA) were used, sampled at 2~kHz to record muscle activity. These sensors were placed on four muscles bilaterally: gastrocnemius lateralis (GS), tibialis anterior (TA), erector spinae (ES), and biceps brachii (BB), following SENIAM guidelines for electrode placement\footnote{www.seniam.org}. Even though the task was symmetric, EMG sensors were placed bilaterally to encounter potential asymmetries in muscle activation that can occur between right and left sides of the body during symmetric tasks, due to mechanisms such as limb dominance~\cite{Mezaour2009Symmetrical}.

The skin surface was prepared by abrading the area with sandpaper and then wiping with alcohol before attaching the EMG sensors. The GS and TA are key in balance control, facilitating plantarflexion and dorsiflexion of the ankle, respectively \cite{doi:10.3233/WOR-131735, app14114635}. The ES plays a crucial role in trunk stabilisation and demonstrates a higher activity when handling loads further from the centre of the body \cite{doi:10.3233/WOR-131735, Madinei03062018}. Overexertion of this muscle is associated with fatigue and lower back pain, which makes it essential to monitor, to assess postural stress and balance control \cite{Hu03082015, Bonato2003}. Furthermore, since the ES contributes to trunk inclination, analysing its activity also provides insights to posture. The BB was also monitored as it is one of the primary muscles involved in lifting tasks. While not directly interacting with the tail or supporting its weight, its activity is recorded to look for indications if the robotic tail  affects the upper limbs in response to performing a task with a new wearable robot.  To evaluate muscle activity, EMG signals were normalised against maximum voluntary isometric contraction (MVIC). Before the experiment, the participants performed an MVIC task for each muscle recorded. The raw EMG signal is demeaned, then filtered using a 4th-order high-pass Butterworth filter (10~Hz cut-off). The signal is then rectified, and a 4th-order low-pass Butterworth filter (1~kHz cut-off) is applied to generate the linear envelope. For all muscles, to remove the potential for muscle activation asymmetries between the left and rights sides of the body, their corresponding means were taken to give a combined average value.

\subsection*{Statistical analysis}

We used \textsc{Matlab} (MathWorks, USA) to extract data for the load placement and load retrieval tasks using software-based timing to standardise start and end points for all repetitions. In a small number of trials ($<3\%$), participants made errors such as colliding with the platform during load movement or not grasping the load properly, leading to mid-trial pauses and re-adjustments. To prevent additional muscle fatigue, these trials were not repeated during the experiment. Instead, they were identified and excluded from the analysis. To statistically evaluate the significance of differences in the group level data for all measurement techniques, a repeated measures analysis of variance (RANOVA) test was performed to assess the significances between conditions. This was followed by a Mauchly's test to check for sphericity of data. In the case of a violation of sphericity, a Greenhouse-Geisser (GG) correction is applied to the RANOVA. Post-hoc comparisons were performed using a Tukey test if sphericity was validated, or a Bonferroni test if not. Statistical significance was set at $\alpha  = 0.05$.

\section*{Data availability:}
The data that support the findings of this study are provided in the supplementary materials Data S1. The time-series data of the individual participants are are available from the corresponding author upon reasonable request.

\section*{Acknowledgements} 
This study was funded by UK Research and Innovation EPSRC grant EP/T027746/1. The funder played no role in study design, data collection, analysis and interpretation of data, or the writing of this manuscript. We would like to thank all the participants for their time in completing our study. We also thank the Queen Mary Ethics Committee for their support in gaining approval to conduct this study.

\section*{Author Contributions}
Eisa Anwar built the wearable robotic tail, designed and conducted the associated experiment, performed the statistical analysis, and wrote the manuscript. Sajeeva Abeywardena contributed to the design parameters of the tail and the experimental setup. Stuart Miller provided guidance on data collection and supported statistical analysis. Ildar Farkhatdinov provided feedback throughout the process, offering insights on the experimental design, and informed the writing and layout of the paper.

\section*{Competing Interests} 
All authors declare that there are no competing interests.

\printbibliography %

\end{document}